%% file: acl_latex.tex
\documentclass[11pt]{article}

\usepackage[]{acl}

\usepackage{times}
\usepackage{latexsym}

\usepackage[T1]{fontenc}

\usepackage[utf8]{inputenc}

\usepackage{microtype}

\usepackage{inconsolata}

\usepackage{graphicx}
\usepackage{subcaption}
\usepackage{booktabs}
\usepackage{multirow}
\usepackage{makecell}
\usepackage{worldflags}
\usepackage{arydshln}

\usepackage[most]{tcolorbox}
\usepackage{enumitem}

\newtcolorbox[auto counter, number within=subsection]{promptbox}[2][]{%
    colback=gray!5!white,      
    colframe=gray!75!black,    
    title={\textbf{Prompt \thetcbcounter:} #2}, 
    fonttitle=\bfseries\sffamily,
    fontupper=\small\ttfamily, 
    breakable,                 
    enhanced,                  
    sharp corners,             
    boxrule=0.8pt,             
    #1                         
}

\newcommand{\dataset}{\textsc{DeepSEA}}
\def\ztitle{\dataset{}: A Multilingual Evaluation Benchmark for Southeast Asia}\title{\ztitle}

\author{
  Jiuzhou Han\thanks{Work done during an internship at Google.} \hspace{1.2em}
  Abhishek Bijay Mishra \hspace{1.2em}
  Trevor Cohn \hspace{1.2em}
  Shubham Mittal \\
  Google Research \\
  \texttt{jiuzhou.han@monash.edu}, \texttt{\{abhishekm, tcohn, shmittal\}@google.com}
}

\begin{document}
\maketitle
\begin{abstract}
\input{Sections/Abstract}
\end{abstract}

\section{Introduction}
\input{Sections/Introduction}

\section{Related Work}
\input{Sections/Related}

\section{\dataset{}}
\input{Sections/Dataset}

\section{Experiments}
\input{Sections/Experiments}

\section{Analysis}
\input{Sections/Analysis}

\section{Conclusion}
\input{Sections/Conclusion}

\section*{Limitations}
\input{Sections/Limitations}

\section*{Acknowledgments}
\input{Sections/Ack}

\bibliography{anthology_0,custom}

\appendix
\input{Sections/Appendix}
\label{sec:appendix}

\end{document}

%% file: Sections/Abstract.tex
While language models have demonstrated remarkable capabilities, their performance remains disproportionately skewed towards high-resource languages. 
This disparity is particularly acute for the 700 million inhabitants of Southeast Asia, where vast linguistic diversity remains under-represented in existing benchmarks, often suffering from limited language coverage or narrow task scopes.
To address this, we introduce \dataset{}, a comprehensive benchmark constructed from established datasets through professional human translation into 12 Southeast Asian languages, spanning both high-resource (e.g., Indonesian, Thai, Vietnamese) and low-resource (e.g., Burmese, Khmer, Lao) tiers across five distinct tasks: mathematical reasoning, scientific reasoning, question answering, abstractive summarization, and machine translation. 
Extensive evaluation of frontier proprietary and open-weights models reveals that while frontier models like Gemini-3.7-Flash maintain robustness across resource tiers, smaller models face severe performance degradation on low-resource languages. 
Crucially, models perform better in reasoning and summarization when pivoted through English rather than target languages, exposing a critical dependency on English representations, especially for low-resource languages.
Beyond these observations, we provide a cross-lingual error analysis and a cross-script tokenizer fertility analysis.
We hope these findings shed light on the development of next-generation inclusive multilingual models.\footnote{\dataset{} is open source and available at \mbox{\url{https://github.com/google-research/deep-sea}}.}

%% file: Sections/Introduction.tex
\begin{figure*}[ht]
    \centering
    \includegraphics[width=\textwidth]{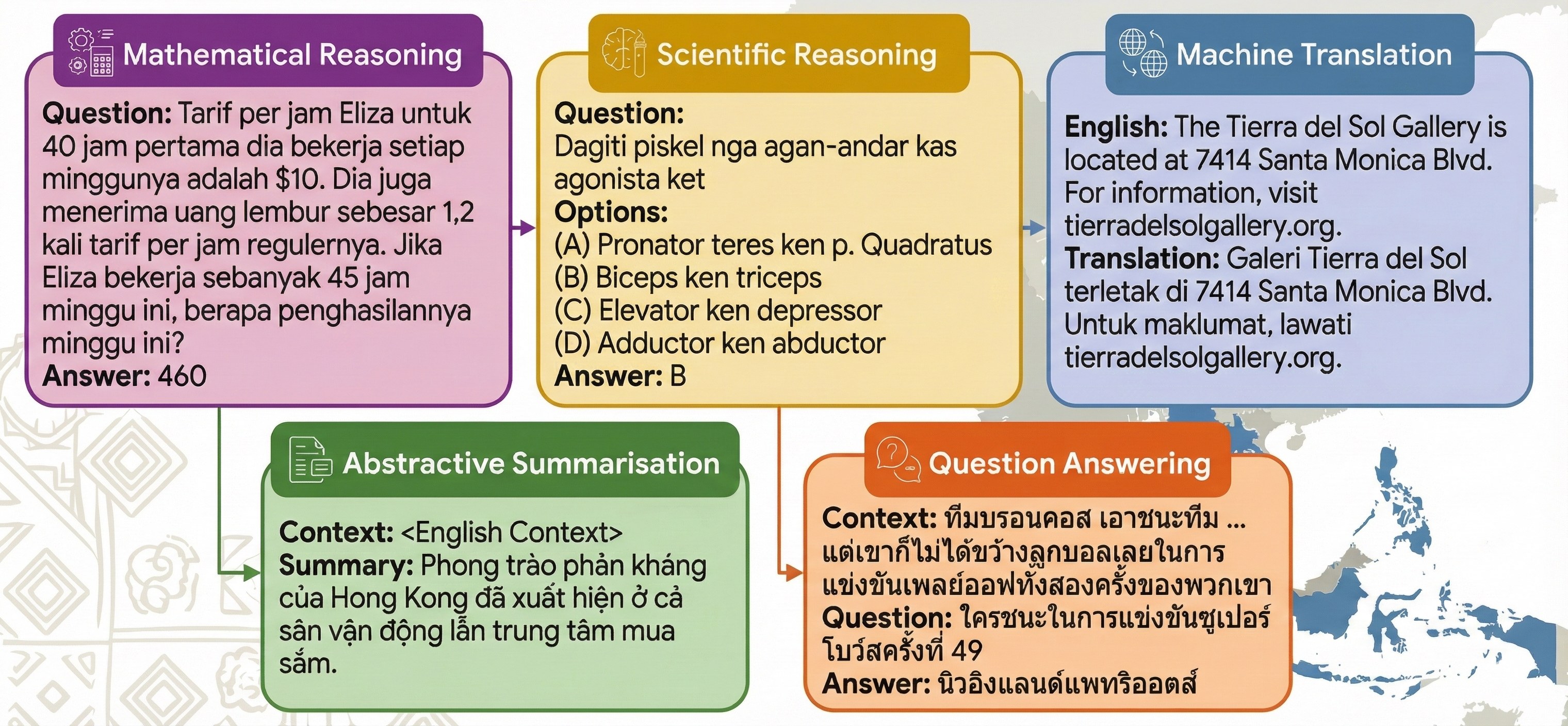}
    \caption{Demonstration of various tasks covered in \dataset{}, illustrating specific examples of inputs and outputs.}
    \label{fig:tasks}
\end{figure*}

The advent of Large Language Models (LLMs) has revolutionized natural language processing, demonstrating remarkable capabilities in reasoning, generation, and understanding~\cite{DBLP:journals/corr/abs-2303-08774, DBLP:journals/corr/abs-2507-06261, DBLP:journals/corr/abs-2501-12948}. However, despite these advancements, the benefits of LLMs remain disproportionately skewed towards high-resource languages, particularly English and European languages. This disparity creates a barrier for the 700 million inhabitants of Southeast Asia (SEA), a region characterized by high linguistic diversity and a complex tapestry of high, medium, and extremely low-resource languages~\cite{DBLP:conf/acl/SusantoHMNYLRL025, lovenia-etal-2024-seacrowd}.

While recent initiatives have evaluated multilingual LLMs across the SEA region, existing benchmarks often suffer from limited language coverage or narrow task scopes. Leading benchmarks like SeaEval~\cite{wang-etal-2024-seaeval}, SeaExam, and SeaBench~\cite{liu-etal-2025-seaexam-new} introduce valuable frameworks for cultural reasoning and local examinations, yet cluster disproportionately around major national languages (primarily Indonesian, Vietnamese, and Thai). SEA-HELM~\cite{DBLP:conf/acl/SusantoHMNYLRL025} introduces a holistic five-pillar framework but evaluates only five languages, leaving vast portions of the region unrepresented. Furthermore, while SEACrowd~\cite{lovenia-etal-2024-seacrowd} consolidates resources for nearly 1,000 languages, it serves primarily as a data hub rather than a targeted evaluation suite across the resource spectrum.

A critical gap persists in the lack of a unified evaluation benchmark covering both major national languages and underserved low-resource regional languages. To bridge this gap, we introduce \textbf{\dataset{}}, the first comprehensive, human-translated benchmark designed to rigorously evaluate LLMs across 12 Southeast Asian languages. \dataset{} assesses capabilities in understanding, multi-step reasoning, and generation across five distinct tasks: mathematical reasoning (MGSM), scientific reasoning (IndoMMLU), question answering (XQuAD), abstractive summarization (CrossSum), and machine translation (WMT24++).

\pagebreak
Our contributions are as follows:
\begin{itemize}
    \item We introduce \dataset{}, a human-translated benchmark spanning 12 SEA languages across high- and low-resource tiers, evaluated on five understanding and generation tasks.
    \item We provide a systematic evaluation of the LLM landscape, ranging from frontier proprietary systems to open-weights models.
    \item We conduct a granular performance analysis across languages, isolating key bottlenecks and providing actionable insights for future AI development for the SEA region.
\end{itemize}

%% file: Sections/Related.tex
\paragraph{High-Coverage Parallel Benchmarks} Multilingual evaluation has traditionally relied on $N$-way parallel corpora for cross-lingual comparison \citep{conneau-etal-2018-xnli}.
The translation evaluation dataset, FLORES-200 \citep{DBLP:journals/corr/abs-2207-04672} and its reading comprehension derivative Belebele \citep{bandarkar-etal-2024-belebele} are the only benchmarks covering all 12 SEA languages examined in this study to the best of our knowledge.
Topic classification datasets SIB-200 \citep{adelani-etal-2024-sib} and Taxi1500 \citep{ma-etal-2025-taxi1500} also cover all 12 languages, but restrict labels to English.
Other high-coverage parallel datasets include NTREX-128 \citep{federmann-etal-2022-ntrex} (translation) and MASSIVE \citep{fitzgerald-etal-2023-massive} (slot filling), both covering 8 of the 12 languages. However, the task diversity in these datasets is limited. 

\paragraph{Global Knowledge and Reasoning Benchmarks} The above parallel datasets are translated from an English base, but recent efforts have shifted to more globally representative benchmarks. Global-MMLU \citep{singh-etal-2025-global} translates only the subset of MMLU that is culture-agnostic.
In contrast, non-parallel benchmarks like MultiLoKo \citep{DBLP:journals/corr/abs-2504-10356}, TyDiQA \citep{clark-etal-2020-tydi}, and M3exam \citep{DBLP:conf/nips/ZhangAGCB23} are built using locally sourced native questions across diverse languages. Similarly, INCLUDE \citep{DBLP:conf/iclr/RomanouFSNSMACH25} and Global-PIQA \citep{chang2025globalpiqaevaluatingphysical} leverage large-scale crowdsourcing to collect high-coverage, culturally grounded items.
While all these global datasets cover a few SEA languages, most of the 12 languages examined in this work are missing from each, with Sundanese, Ilocano, and Cebuano being present in none, despite accounting for 60 million speakers in aggregate.

\paragraph{SEA Regional Benchmarks}
Recent evaluation frameworks for Southeast Asia span both regional and dialect-specific scopes. At the regional level, SeaEval \citep{wang-etal-2024-seaeval}, SeaExam, and SeaBench \citep{liu-etal-2025-seaexam-new} emphasize cultural reasoning and local examinations, while SEA-HELM \citep{DBLP:conf/acl/SusantoHMNYLRL025} introduces a holistic five-pillar framework, though limited to five languages. Significant attention has also been directed towards Indonesia's linguistic diversity; IndoMMLU \citep{koto-etal-2023-large} targets academic proficiency, while LoraxBench \citep{aji-cohn-2025-loraxbench} and NusaX \citep{winata-etal-2023-nusax} provide extensive coverage of local languages through translated and parallel tasks.
Beyond Indonesia, specialized benchmarks target other major national languages, such as \textsc{Batayan} \citep{montalan-etal-2025-batayan} for Filipino, ThaiCLI \citep{kim-etal-2025-representing} for Thai cultural nuances, and ViQuAD \citep{nguyen-etal-2020-vietnamese} for Vietnamese.

%% file: Sections/Dataset.tex
\subsection{Languages of Focus}

The benchmark targets 12 Southeast Asian languages, stratified into two resource categories, high and low. Group high comprises Thai, Indonesian, Vietnamese, and Filipino. Group low includes Cebuano, Ilocano, Javanese, Khmer, Lao, Malay, Sundanese, and Burmese. Table~\ref{tab:lang_details} demonstrates the linguistic characteristics of these languages.

\subsection{Task Coverage}

To ensure a comprehensive evaluation, \dataset{} encompasses five distinct tasks derived from established benchmarks. Where target-language coverage was unavailable, we commissioned professional native translators to translate the original instances into the target Southeast Asian languages. Figure~\ref{fig:tasks} illustrates examples of the five tasks.

\begin{table}[t]
\centering
\newcommand{\flag}[1]{\worldflag[length=0.4cm, width=0.26cm]{#1}\hspace{0.2cm}}

\resizebox{\columnwidth}{!}{
\begin{tabular}{lllll}
\toprule
\textbf{Language} & \textbf{Code} & \textbf{Family} & \textbf{Script} & \textbf{Typology} \\ \midrule
\flag{VN}Vietnamese & vi & Austroasiatic & Latin & Analytic \\
\flag{PH}Filipino & fil & Austronesian & Latin & Agglutinative \\
\flag{ID}Indonesian & id & Austronesian & Latin & Agglutinative \\
\flag{TH}Thai & th & Tai-Kadai & Thai & Analytic \\ 
\hdashline
\flag{MY}Malay & ms & Austronesian & Latin & Agglutinative \\
\flag{PH}Cebuano & ceb & Austronesian & Latin & Agglutinative \\
\flag{ID}Javanese & jv & Austronesian & Latin & Agglutinative \\ 
\flag{ID}Sundanese & su & Austronesian & Latin & Agglutinative \\
\flag{MM}Burmese & my & Sino-Tibetan & Myanmar & Agglutinative \\
\flag{PH}Ilocano & ilo & Austronesian & Latin & Agglutinative \\
\flag{KH}Khmer & km & Austroasiatic & Khmer & Analytic \\
\flag{LA}Lao & lo & Tai-Kadai & Lao & Analytic \\ \bottomrule
\end{tabular}
}
\caption{Linguistic characteristics of the covered languages. The first four languages are classified as high-resource, while the remaining eight are low-resource.}
\label{tab:lang_details}
\end{table}

\paragraph{Mathematical Reasoning}
To evaluate arithmetic reasoning ability, we adopt data from the Multilingual Grade School Math (MGSM)~\cite{DBLP:conf/iclr/ShiSF0SVCTRZ0W23}. MGSM contains grade-school level math problems that require multi-step reasoning to solve. 


\paragraph{Scientific Reasoning}
Our scientific reasoning dataset is derived from IndoMMLU~\cite{koto-etal-2023-large}, a massive multitask language understanding benchmark in Indonesian. We specifically leveraged a subset of questions focusing on STEM domains: Biology, Chemistry, and Physics. 

\paragraph{Question Answering}
For this task, we use XQuAD~\cite{artetxe-etal-2020-cross}, consisting of reading comprehension passages accompanied by questions that require extractive answers. 

\paragraph{Abstractive Summarization}
We assess summarization capabilities using CrossSum~\cite{bhattacharjee-etal-2023-crosssum}, a large-scale cross-lingual summarization corpus. This task challenges models to generate a concise summary in a target language based on the provided source document in English. 

\paragraph{Machine Translation}
The machine translation dataset is constructed based on data from WMT shared tasks~\cite{DBLP:conf/wmt/KocmiABBDFFFGGH24, deutsch-etal-2025-wmt24}. We focus specifically on the translation direction from English to the target languages to assess the models' generative capabilities in low-to-high resource settings. 

\subsection{Data Creation and Statistics}

We aggregated existing publicly available data and extended the benchmark by generating new translations for unsupported languages. We commissioned professional native translators to map the data from the respective source languages to the target SEA languages. For MGSM, XQuAD, WMT, and CrossSum, translators worked directly from the English source text. Since IndoMMLU is originally an Indonesian benchmark, translations for this task were conducted from Indonesian into the target low-resource languages (details in Appendix~\ref{appd:bench_details}).

To preserve the meaning and evaluation format of the original instances, translators were provided with explicit translation guidelines such as translating only the designated text field, while using the original article or context, when available, solely as a reference for disambiguation.
They were asked to provide a single natural translation rather than multiple alternatives and to avoid adding clarifications, redundant information, or explanatory content not present in the source.
Numerical expressions were required to retain the same form as the original input (e.g., words versus numerals), while punctuation, quotation marks, capitalization, and the overall formatting of the source text were preserved as closely as possible.
To ensure high translation quality and ethical labour standards, the entire \dataset{} dataset was translated through professional translation vendors, with all contributors receiving fair, competitive compensation well above local minimum wage guidelines (e.g., \$0.16 per word in MGSM).

\begin{table}[t]
\centering
\resizebox{0.48\textwidth}{!}{%
\begin{tabular}{@{}llcc@{}}
\toprule
\textbf{Task} & \textbf{Eval} & \textbf{\# high} & \textbf{\# low} \\ \midrule
Mathematical Reasoning & MGSM & 1,000 & 2,000 \\
Scientific Reasoning & IndoMMLU & 2,940 & 5,880 \\
Question Answering & XQuAD & 4,760 & 9,520 \\
Abstractive Summarization & CrossSum & 1,558 & 3,569 \\
Machine Translation & WMT24++ & 3,988 & 7,976 \\ \bottomrule
\end{tabular}%
}
\caption{Number of samples in \dataset{} eval sets, grouped by 4 high and 8 low-resource languages.}
\label{tab:statistics}
\end{table}

Quality check of the \dataset{} benchmark confirms the high fidelity and accuracy of the translated datasets (refer to Appendix~\ref{appd:quality_check} for more details). 
Table~\ref{tab:statistics} summarizes the statistics of various datasets. 
Except for CrossSum, all tasks maintain a strict one-to-one parallel alignment with English; in CrossSum, splits for Thai (th), Indonesian (id), Vietnamese (vi), and Burmese (my) lack direct counterparts because matching source articles were unavailable in the original corpus.

%% file: Sections/Experiments.tex
\begin{table*}[ht]
\centering
\resizebox{\textwidth}{!}{%
\begin{tabular}{@{} l *{14}{wc{0.9cm}} @{}}
\toprule
\multirow{2.5}{*}{\textbf{Model}}
& \multicolumn{3}{c}{\textbf{\makecell{Mathematical\\Reasoning}}}
& \multicolumn{3}{c}{\textbf{\makecell{Scientific\\Reasoning}}}
& \multicolumn{3}{c}{\textbf{\makecell{Question\\Answering}}}
& \multicolumn{3}{c}{\textbf{\makecell{Abstractive\\Summarization}}}
& \multicolumn{2}{c}{\textbf{\makecell{Machine\\Translation}}} \\
\cmidrule(lr){2-4}
\cmidrule(lr){5-7}
\cmidrule(lr){8-10}
\cmidrule(lr){11-13}
\cmidrule(lr){14-15}
& \multicolumn{3}{c}{Accuracy $\uparrow$}
& \multicolumn{3}{c}{Accuracy $\uparrow$}
& \multicolumn{3}{c}{F1 $\uparrow$}
& \multicolumn{3}{c}{ChrF $\uparrow$}
& \multicolumn{2}{c}{MetricX-24 $\downarrow$} \\
\cmidrule(lr){2-4}
\cmidrule(lr){5-7}
\cmidrule(lr){8-10}
\cmidrule(lr){11-13}
\cmidrule(lr){14-15}
\textbf{Language}
& en & high & low
& en & high & low
& en & high & low
& en & high & low
& high & low \\
\midrule
Gemini-3.7-Flash
& \textbf{99.2} & \textbf{96.5} & \textbf{96.4}
& \textbf{86.5} & \textbf{86.3} & \textbf{86.1}
& \textbf{87.9} & \textbf{85.2} & \textbf{81.7}
& \textbf{37.4} & \textbf{36.4} & \textbf{36.2}
& \textbf{2.4} & \textbf{3.5} \\
Gemini-3-Pro
& 98.0 & 95.6 & 95.4
& 85.0 & 85.3 & 84.3
& 85.6 & 81.2 & 77.6
& 37.0 & 35.9 & 35.4
& 2.5 & 3.7 \\
Gemini-3-Flash
& 98.8 & 95.7 & 95.7
& 85.7 & 85.2 & 84.5
& 87.2 & 83.6 & 79.5
& 36.9 & 36.2 & \textbf{36.2}
& 2.5 & 3.7 \\
GPT-5.6-Luna
& 96.4 & 93.3 & 92.8
& 85.3 & 84.5 & 83.3
& 73.8 & 68.1 & 65.8
& 33.7 & 33.8 & 33.2
& \textbf{2.4} & \textbf{3.5} \\
Claude-Sonnet-5
& 98.8 & 96.4 & 95.4
& 84.9 & 84.6 & 82.6
& 87.1 & 84.3 & 80.3
& 35.7 & 35.8 & 34.0
& 2.6 & 3.8 \\
\midrule
Gemma-3-27B
& \underline{96.0} & 91.7 & 83.6
& \underline{79.2} & \underline{76.2} & \underline{71.5}
& 85.5 & 79.9 & 73.9
& 33.1 & \underline{33.5} & \underline{29.7}
& \underline{2.9} & 6.1 \\
Gemma-3-12B
& 92.8 & 89.7 & 77.2
& 71.0 & 68.5 & 63.6
& 84.7 & 77.3 & 71.5
& 33.4 & 32.5 & 27.5
& 3.1 & 8.6 \\
Qwen2.5-7B-Inst
& 93.2 & 78.3 & 36.2
& 68.2 & 61.3 & 49.2
& 71.9 & 48.5 & 37.9
& 32.9 & 28.4 & 22.2
& 8.2 & 17.6 \\
Llama-3.1-8B-Inst
& 87.6 & 67.3 & 38.8
& 52.0 & 42.3 & 24.5
& 71.1 & 62.9 & 37.2
& \underline{34.8} & 30.3 & 21.5
& 5.9 & 14.8 \\
\hdashline
SEA-LION-v4-27B
& \underline{96.0} & \underline{92.7} & \underline{83.8}
& 78.1 & \underline{76.2} & \underline{71.5}
& \underline{85.6} & \underline{80.1} & \underline{74.4}
& 33.0 & 33.0 & 28.8
& 3.0 & \underline{6.0} \\
Sailor-2-20B
& 92.8 & 87.7 & 77.6
& 51.3 & 37.9 & 34.3
& 74.7 & 57.4 & 46.1
& 31.4 & 30.4 & 27.2
& 4.6 & 12.6 \\
\bottomrule
\end{tabular}
}
\caption{Performance of selected models across different tasks, with languages split among English, high- and low-resource groups as defined in Table~\ref{tab:lang_details}. \textbf{Bold} values indicate the best overall performance, while \underline{underlined} values indicate the best performance among open-weights models.}
\label{tab:baseline_performance}
\end{table*}

\subsection{Models}
We evaluate a diverse spectrum of models across three categories.
Firstly, the frontier proprietary systems comprising Gemini-3.7-Flash \citep{googledeepmind2026gemini37flash}, Gemini-3-Pro \citep{googledeepmind2025gemini3pro}, Gemini-3-Flash \citep{googledeepmind2025gemini3flash}, GPT-5.6-Luna \citep{openai2026gpt56}, and Claude-Sonnet-5 \citep{anthropic2026claudesonnet5}.
Secondly, the leading open-weights models, including Gemma-3 \citep{DBLP:journals/corr/abs-2503-19786}, Qwen2.5 \citep{DBLP:journals/corr/abs-2412-15115}, and Llama-3.1 \citep{DBLP:journals/corr/abs-2407-21783}.
Lastly, we also evaluate the regionally adapted models specifically optimised for Southeast Asian languages, namely SEA-LION \citep{DBLP:conf/ijcnlp/NgNHTLLYNSCRLHTTSTTOOMC25} and Sailor2 \citep{DBLP:journals/corr/abs-2502-12982}.
Further, additional model results are provided in Appendix Table~\ref{tab:new_model_comparison}.

\subsection{Inference Setup}
All models are evaluated in a zero-shot setting with greedy decoding (temperature $=0$) to ensure deterministic results.
For reasoning-intensive tasks (i.e., mathematical and scientific reasoning), we explicitly instruct models to generate step-by-step reasoning.
To maintain linguistic consistency, models are directed to reason in the same language in which the problem is presented.
Complete prompt templates for each task are provided in Appendix~\ref{appd:prompts}.

\subsection{Evaluation Metrics}

We employ task-specific metrics across the \dataset{} benchmark.
For reasoning-intensive tasks (mathematical and scientific reasoning), we report standard accuracy, and for question answering, we measure token-level F1.
For abstractive summarization, we report chrF \citep{DBLP:conf/wmt/Popovic15}, following prior multilingual benchmarks \citep{montalan-etal-2025-batayan}.
Finally, machine translation performance is assessed using MetricX-24 \citep{juraska-etal-2024-metricx}. Table~\ref{tab:baseline_performance} summarizes the results across all evaluated models and tasks.

\subsection{Results}
\label{results}
\subsubsection{Results Across Models}

\paragraph{Frontier Proprietary Models}
Gemini-3.7-Flash consistently achieves state-of-the-art performance across all evaluated tasks and demonstrates particular robustness in maintaining high accuracy across language tiers, exhibiting minimal degradation from English to high- and low-resource languages. For instance, in mathematical reasoning (MGSM), Gemini-3.7-Flash maintains a score of 96.4 in the low-resource category, significantly outperforming the best open-weights model, SEA-LION-v4-27B, which scores 83.8.
Claude-Sonnet-5 is also highly competitive, particularly in mathematical reasoning and question answering, while GPT-5.6-Luna also performs strongly on the reasoning tasks and ties Gemini-3.7-Flash for the best machine translation performance.

\paragraph{Open-Weights Models}
Among general-purpose open-weights models, we observe a clear scaling trajectory where increased model capacity substantially strengthens cross-lingual transfer.
Gemma-3-27B consistently outperforms Gemma-3-12B across tasks, achieving 71.5 accuracy on low-resource scientific reasoning compared to 63.6 for the 12B variant.
In contrast, the smaller generalist models (Qwen2.5-7B-Inst and Llama-3.1-8B-Inst) suffer from a severe performance collapse on low-resource languages despite respectable English accuracy.
In mathematical reasoning, Qwen2.5-7B-Inst drops from 93.2 in English to 78.3 (high) and 36.2 (low)—a degradation of over 60\%—whereas the frontier proprietary models or open-weights regional models maintain much flatter cross-lingual curves.
This highlights that the small language models lag behind in their multilingual capabilities, unless specialised for the target languages.

\paragraph{Regional Models}
The performance of models specialised for SEA region, such as SEA-LION-v4-27B, shows impressive competitiveness against the generalist models of similar size, particularly in high-resource languages. It effectively matches or surpasses Gemma-3-27B in XQuAD and WMT24++, suggesting that regionally pre-trained models remain a vital strategy for optimising performance in specific linguistic geographies.

\begin{figure*}[t]
    \centering
    \includegraphics[width=\textwidth]{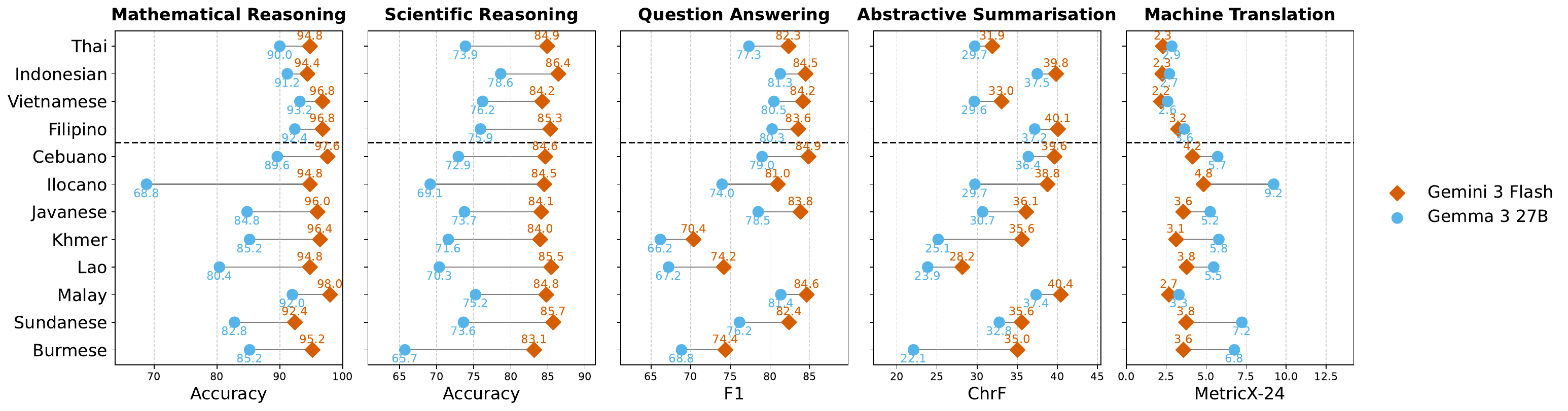}    
    \caption{Frontier models maintain robustness across resource tiers, while open-weights models show severe degradation on low-resource languages. Gemini-3-Flash and Gemma-3-27B are denoted by orange diamonds and blue circles, respectively. The dotted line separates high- and low-resource languages.}
    \label{fig:gemini_gemma_comparison}
\end{figure*}

\subsubsection{Results Across Tasks}

\paragraph{Reasoning Tasks (MGSM and IndoMMLU)}

Mathematical and scientific reasoning appear to be the most discriminative tasks for evaluating cross-lingual capability. While English performance is relatively high across most models, the gap widens considerably for low-resource languages. Gemini-3.7-Flash demonstrates particularly strong robustness, decreasing only from 99.2 to 96.4 on Mathematical Reasoning and from 86.5 to 86.1 on Scientific Reasoning between English and low-resource languages. Among general-purpose open-weights models, Gemma-3-27B achieves the strongest low-resource performance, with scores of 83.6 and 71.5 on the two reasoning tasks, respectively. This suggests that although reasoning ability may transfer across languages, linguistic accessibility remains a major bottleneck for weaker models.

\paragraph{Question Answering (XQuAD)}

Performance on XQuAD exhibits a sharper cross-lingual drop between English and SEA languages than the reasoning tasks. Gemini-3.7-Flash maintains the strongest performance, decreasing from 87.9 in English to 81.7 on low-resource languages. Notably, GPT-5.6-Luna performs substantially lower on question answering relative to other tasks, achieving F1 scores of 73.8, 68.1, and 65.8 across English, high-resource, and low-resource tiers, respectively.
Our error analysis reveals that this drop is driven by poor instruction-following: while the prompt strictly requires answer substring extraction from the context, the model often outputs paraphrased responses and gets penalised by token-level F1.


\paragraph{Abstractive Summarization (CrossSum)}

The CrossSum task reveals relatively stable performance across language tiers for the strongest Gemini models. Gemini-3.7-Flash achieves chrF scores of 37.4, 36.4, and 36.2 for English, high-resource, and low-resource languages, respectively, while Gemini-3-Flash also achieves 36.2 chrF on low-resource languages. However, the low-resource tier remains a significant bottleneck for smaller generalist models, with Qwen2.5-7B-Inst and Llama-3.1-8B-Inst dropping to 22.2 and 21.5, respectively. Among general-purpose open-weights models, Gemma-3-27B achieves the strongest low-resource result of 29.7 chrF, slightly outperforming the regional SEA-LION-v4-27B at 28.8 chrF.

\paragraph{Machine Translation (WMT24++)}

Gemini-3.7-Flash and GPT-5.6-Luna show the best performance, achieving MetricX-24 scores of 2.4 and 3.5 for high- and low-resource languages, respectively.
Among open-weights models, Gemma-3-27B achieves the strongest high-resource score of 2.9 and remains competitive with SEA-LION-v4-27B on low-resource languages with a score of 6.1.
Smaller models struggle significantly more, with Qwen2.5-7B-Inst and Llama-3.1-8B-Inst reaching scores of 17.6 and 14.8, resp., in low-resource languages. This reveals that translation quality for under-represented languages strongly differentiates the frontier and open-weights models.




\subsubsection{Results across Languages}
Figure~\ref{fig:gemini_gemma_comparison} shows the scores across 12 SEA languages, using Gemini-3-Flash and Gemma-3-27B as the representative frontier and open-weights systems.

\paragraph{Near-Parity in High-Resource Languages}
Both the models perform consistently well across the high-resource languages, maintaining high scores on all five tasks. 
Across all 12 languages, Gemini-3-Flash nearly achieves cross-lingual parity, maintaining $>92\%$ in math and $>83\%$ in science.
In contrast, Gemma-3-27B fluctuates widely, suffering steep drops on low-resource languages, showing that cross-lingual parity remains a key strength of the frontier models.

\paragraph{Disparity in Low-Resource Languages}
Low-resource languages show substantial performance degradation, especially in non-Latin scripts: Burmese, Khmer, and Lao.
These three languages yield the lowest scores in question answering ($<75$ F1) and summarization (e.g., $22.1$ chrF for Burmese in Gemma-3-27B).
Regional Latin-script languages like Ilocano also suffer sharp drops ($68.8\%$ in math and $9.2$ in translation), revealing that data scarcity and script differences remain major hurdles for open-weights models.

%% file: Sections/Analysis.tex
\begin{figure*}[t]
    \centering
    \includegraphics[width=\textwidth]{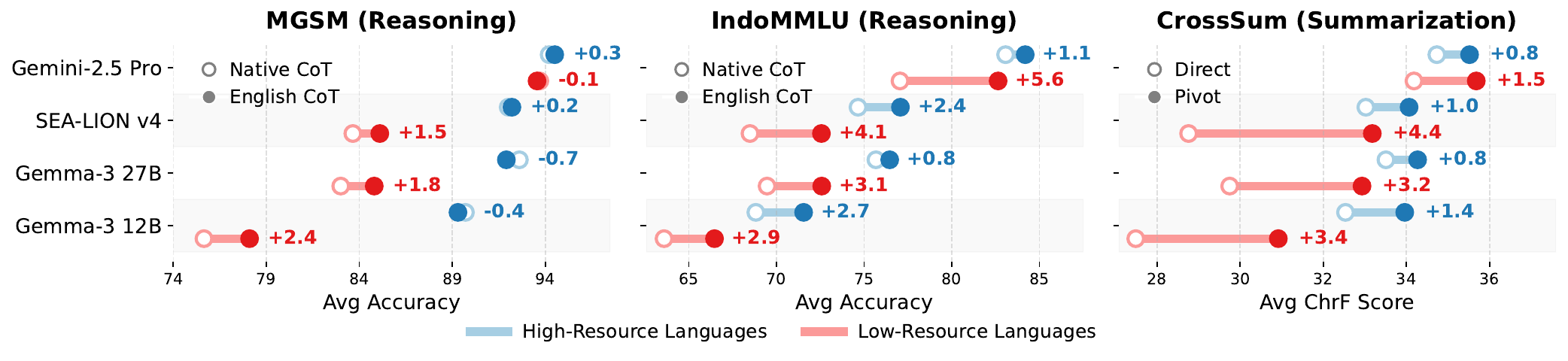}
    
    \caption{Reasoning in English yields consistently higher accuracy than native-language reasoning, especially for low-resource languages. The two left panels compare English and native CoT on reasoning tasks, while the right panel compares direct and English-pivot summarisation.}
    \label{fig:native_en_cot_comparison}
\end{figure*}

\subsection{Language of Reasoning} 

To investigate the effect of language in reasoning-intensive tasks, we evaluate models on MGSM and IndoMMLU using two prompting methodologies.
In the \textit{English CoT} method, models receive an English-language prompt that explicitly instructed them to provide their step-by-step reasoning in English.
Whereas in the \textit{native CoT} method, the entire prompt is translated into the language of each test instance, and models are explicitly instructed to provide their reasoning in that language (e.g., in Thai for Thai instances). 
These prompting methods differ from the default task prompts used in Section~\ref{results}, which have the general instruction to reason in the same language as the problem.
For brevity, Appendix~\ref{appd:prompts} presents only the English versions of the prompt templates for the mathematical and scientific reasoning tasks as representative examples; the corresponding native CoT prompts were fully translated into each target language. The results are provided in Figure~\ref{fig:native_en_cot_comparison}.

\paragraph{Improvements using English CoT}
Across both the benchmarks, models using the English CoT method consistently achieve better or equal accuracy than the native CoT prompt.
This shows that models often perform better when reasoning instructions are provided in English and they are explicitly asked to reason in English.
While English CoT prompting may help mitigate linguistic bottlenecks associated with target-language prompting, it does reveal that models are less capable to perform when asked in languages other than English.

\paragraph{Performance Gradient in Native CoT}
Performance gap between native and English CoT seems to exacerbate from high- to low-resource languages.
In IndoMMLU, Gemma-3-12B and SEA-LION-v4 get performance jump when using English CoT.
This further supports earlier observation that the models are less likely to perform when natively thinking in low-resource languages.

\paragraph{Task Sensitivity}
The benefit of English CoT appears more distinct in scientific reasoning (IndoMMLU) than in mathematical reasoning (MGSM). 
In MGSM, the performance bars for native and English CoT are often closer in height, possibly because MGSM is an easier task than IndoMMLU.
This may also suggest that mathematical reasoning is less sensitive to the choice of CoT prompting language, whereas scientific reasoning appears more sensitive to the difference between English and native CoT prompting.

\paragraph{Frontier Model Robustness}
Gemini-2.5-Pro exhibits the highest degree of stability between reasoning strategies. In MGSM, its Native CoT performance is nearly identical to its English CoT performance across both high and low tiers. This implies that frontier models possess strong multilingual reasoning capabilities, reducing the necessity of an English pivot compared to open-weights models.

\subsection{Summarisation Strategy}
To evaluate the efficacy of cross-lingual information abstraction, we conduct experiments on the CrossSum dataset using two prompting strategies.
The first, ``Direct'', instructs the model to generate the summary in the target language directly from the English source.
The second, ``Pivot-based'', explicitly decomposes the task: first summarising the English source into an English summary and then translating that summary into the target language. The prompt templates are given in Appendix~\ref{appd:prompts}.

As shown in the results in Figure~\ref{fig:native_en_cot_comparison}, the benefits of the Pivot-based decomposition are clearly evident for low-resource languages.
The performance delta is consistently wider when compared to high-resource languages.
For instance, SEA-LION-v4-27B shows a large +4.4 chrF boost for low-resource languages whereas high-resource languages get +1.0 lift.
This indicates that while models may possess sufficient internal knowledge to summarise directly into high-resource languages, their abstractive capabilities in low-resource settings remain brittle.
The Pivot strategy effectively converts a complex abstractive task into a simpler translation task, which current models handle with greater robustness.

We attribute these gains to the effective separation of information compression (summarisation) from linguistic production (translation). By leveraging English as a high-fidelity intermediate representation, the Pivot approach secures the semantic core before addressing the perplexity challenges of the target language. This strategy mitigates the task interference typical of the Direct setting, where the dual burden of summarising and translating simultaneously often leads to content loss or hallucinations in low-resource languages.

\subsection{Error Analysis}

Beyond metrics, we conducted a granular analysis of failure modes in low-resource languages. We focus on two critical phenomena: Cross-Lingual Lexical Interference in reasoning tasks and Contextual Hallucinations in extractive QA tasks.

\paragraph{Cross-Lingual Interference}  While investigating incorrect responses in the mathematical reasoning task, we identified a specific class of errors driven by cross-lingual lexical interference. This occurs when the model correctly identifies a lexical token but resolves its semantics based on a related, often more dominant, language within the same family.

An interesting example is observed in Ilocano. In the dataset, the term ``lawas'' is used to denote ``week''. However, in Cebuano and other Visayan languages, the same lexeme ``lawas'' signifies ``body''. In several failed Ilocano reasoning chains, we observed models generating contexts related to biological or physical health instead of temporal ones, suggesting the model conflated the Ilocano prompt with Cebuano semantics.

Although this specific error type is not the dominant failure mode, its presence in low-resource settings highlights a critical ambiguity issue in multilingual models. We posit that this interference risk may also exist for other linguistically proximate pairs in the region, such as Thai and Lao (Tai-Kadai family) or Indonesian and Malay (Austronesian family), where shared vocabulary often possesses divergent definitions. While likely less frequent in high-resource pairs due to stronger language identification capabilities, this semantic overlap presents a latent challenge for zero-shot inference in under-represented languages like Lao or Ilocano, where the model's internal representation may be biased towards a dominant, closely related language.

\begin{figure}[t]
    \centering
    \includegraphics[width=0.8\linewidth]{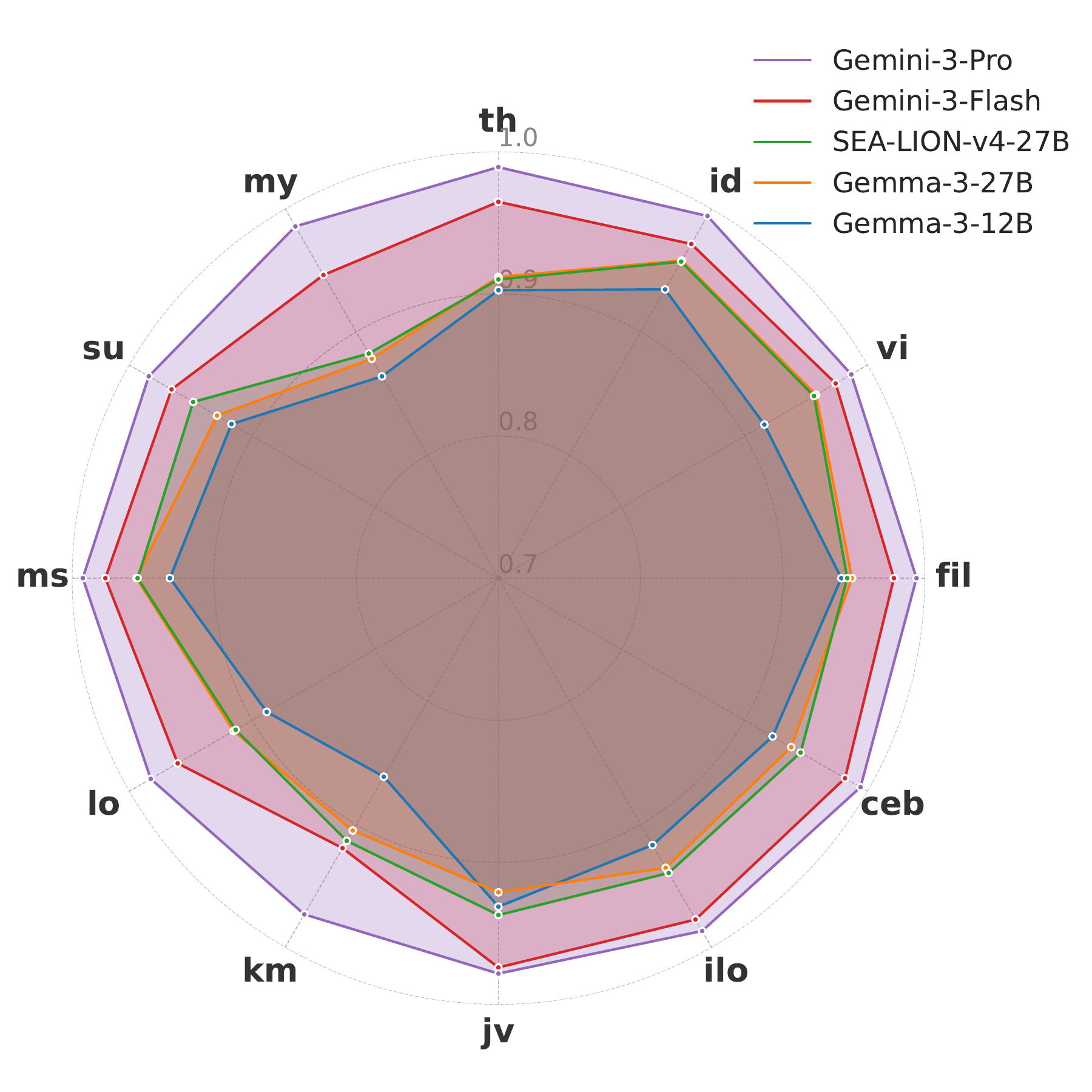}
    \caption{Radar chart illustrating the answer presence rate across 12 SEA languages in XQuAD dataset.}
    \label{fig:xquad_presence_rate}
\end{figure}

\paragraph{Contextual Hallucination} In extractive question answering, models are instructed and constrained to extract answer spans directly from the given context.
We define \textit{contextual hallucination} as generating an answer that does not exist as a substring in the source text, distinguishing span selection errors from complete grounding failures (where entire context is ignored).
To quantify this, we calculate the \textit{answer presence rate}, the percentage of generated answers found verbatim in the context.

As shown in Figure~\ref{fig:xquad_presence_rate}, Gemini-3-Pro outperforms other models by significant margin, indicating its higher instruction following capabilities. 
While most of the SEA languages remain stable, Burmese (my) and Khmer (km) exhibit a significant \emph{grounding collapse} across the other four models.
This degradation creates a sharp divergence along script lines. 
Latin-based languages usually maintain a high answer presence rate, with models consistently adhering to the source text.
In contrast, languages utilising complex abugida scripts (Burmese, Khmer, Lao, and Thai) show a marked reduction in grounding, especially for open-weights models.
Because these languages possess native numeral systems, models frequently generate Western Arabic digits even when the context text uses traditional script-specific numerals.

\begin{figure}[t]
    \centering
    \includegraphics[width=\linewidth]{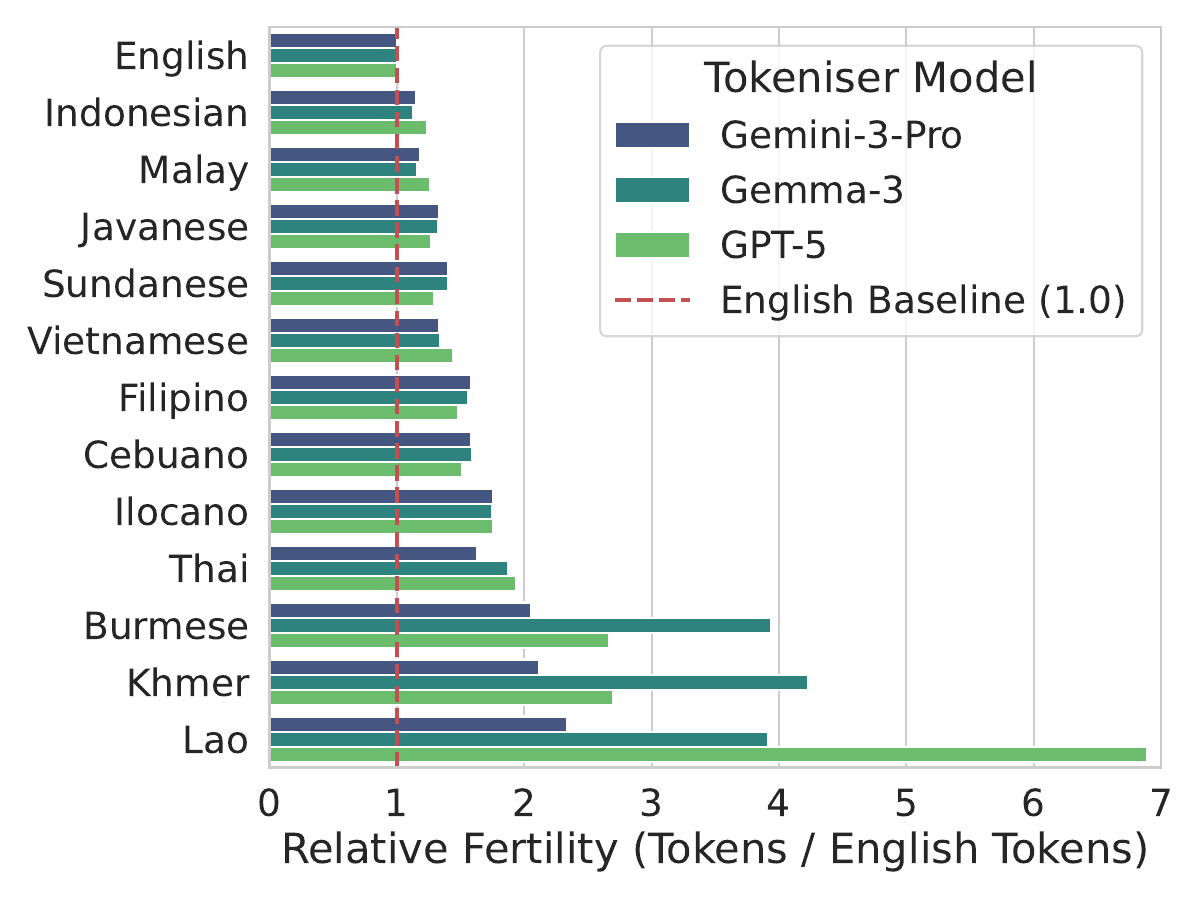}
    \caption{Relative token fertility across SEA languages when compared to English. Values represent the ratio of tokenized tokens in the given language to English tokens (baseline 1.0), calculated using three tokenizers. Higher values indicate lower tokenization efficiency.}
    \label{fig:fertility}
\end{figure}

\subsection{Tokenizer Analysis}

In Figure~\ref{fig:fertility}, we compare the relative token fertility across SEA languages using Gemini-3-Pro, Gemma-3 and GPT-5 tokenizers. We observe a shared pattern of high efficiency for Latin-script languages (e.g., Indonesian, Malay, Vietnamese), where all models align closely with the English baseline. However, a distinct performance gap emerges in non-Latin scripts. While all models exhibit increased fertility for languages such as Burmese, Khmer, and Lao, Gemini-3-Pro and Gemma-3 consistently demonstrate superior compression compared to GPT-5. This disparity is most pronounced in Lao, where GPT-5 suffers from extreme fragmentation with a fertility rate approaching 7.0, whereas Gemini-3-Pro has 2.3. These results indicate that while Gemini-3-Pro and Gemma-3 maintain reasonable efficiency across scripts, GPT-5's tokenisation strategy may incur disproportionately high computational costs and latency for non-Latin Southeast Asian languages.

%% file: Sections/Conclusion.tex
We introduced a novel multilingual benchmark, \dataset{}, designed to evaluate language models across 12 SEA languages, spanning both high- and low-resource tiers. 
Our extensive evaluation of frontier proprietary models and open-weights models establishes strong baselines on this benchmark and reveals a significant performance disparity among low-resource languages.
Further, we identified a critical reliance on English representations, where models consistently perform better when utilising English as an intermediate step rather than operating directly in the target language. 
We also uncovered distinct failure modes in low-resource settings, such as cross-lingual lexical interference and grounding collapse in abugida scripts. 
We hope these findings, alongside the release of the \dataset{} benchmark, will drive the development of next-generation models, which are more inclusive of the SEA region.

%% file: Sections/Limitations.tex
While \dataset{} provides a robust benchmark for evaluating SEA languages, it faces two primary limitations.
Firstly, our dataset construction primarily relies on translating English-centric (or Indonesian-centric for IndoMMLU) benchmarks into target languages. 
Although we employed professional translators and a strict quality control process, this approach may inadvertently introduce translation artifacts, potentially favouring models that have been trained heavily on translated data rather than native text.
Secondly, despite our efforts to span diverse resource tiers, our coverage is limited to 12 distinct languages. Southeast Asia is a region of immense linguistic complexity, hosting hundreds of dialects and extremely low-resource languages that fall outside the scope of this work.

%% file: Sections/Ack.tex
We thank Lucas Bandarkar, Honglin Yu and the multilinguality team in Google Research Australia for insightful discussions on this work. We are grateful to Dinesh Tewari for helping in working with the professional translation vendors. We would also like to thank Grace Chung for her constant guidance and unwavering support.

%% file: Sections/Appendix.tex
\twocolumn[\centering \Large\bfseries \ztitle\\ \large (Appendix) \\ \vspace{2ex} ]

\section{\dataset{} Construction Details}
\label{appd:bench_details}
We distinguish the existing publicly available languages from the additional human translations created for \dataset{} construction in Table~\ref{tab:dataset-availability}.
Further, the detailed data statistics are provided in Table~\ref{tab:statistics_all}.

\section{\dataset{} Quality Check}
\label{appd:quality_check}

\begin{figure}[t]
    \centering
    \begin{minipage}[c]{0.25\textwidth}
        \centering
        \resizebox{\linewidth}{!}{%
            \begin{tabular}{@{}lcc@{}}
                \toprule
                \textbf{Task} & \textbf{Auto-MQM} & \textbf{Human Eval.} \\
                \midrule
                MGSM & 0.86 & 90.75\% \\
                CrossSum & 0.80 & 95.93\% \\
                XQuAD & 0.80 & 95.11\% \\
                WMT24++ & 0.88 & - \\
                IndoMMLU & 0.95 & - \\
                \midrule
                \textbf{Average} & \textbf{0.86} & \textbf{93.93\%} \\
                \bottomrule
            \end{tabular}%
        }
        \label{tab:task_quality_scores}
    \end{minipage}
    \hfill 
    \begin{minipage}[c]{0.22\textwidth}
        \centering
        \includegraphics[trim={4cm 3.3cm 3.3cm 3.3cm}, clip, width=\linewidth]{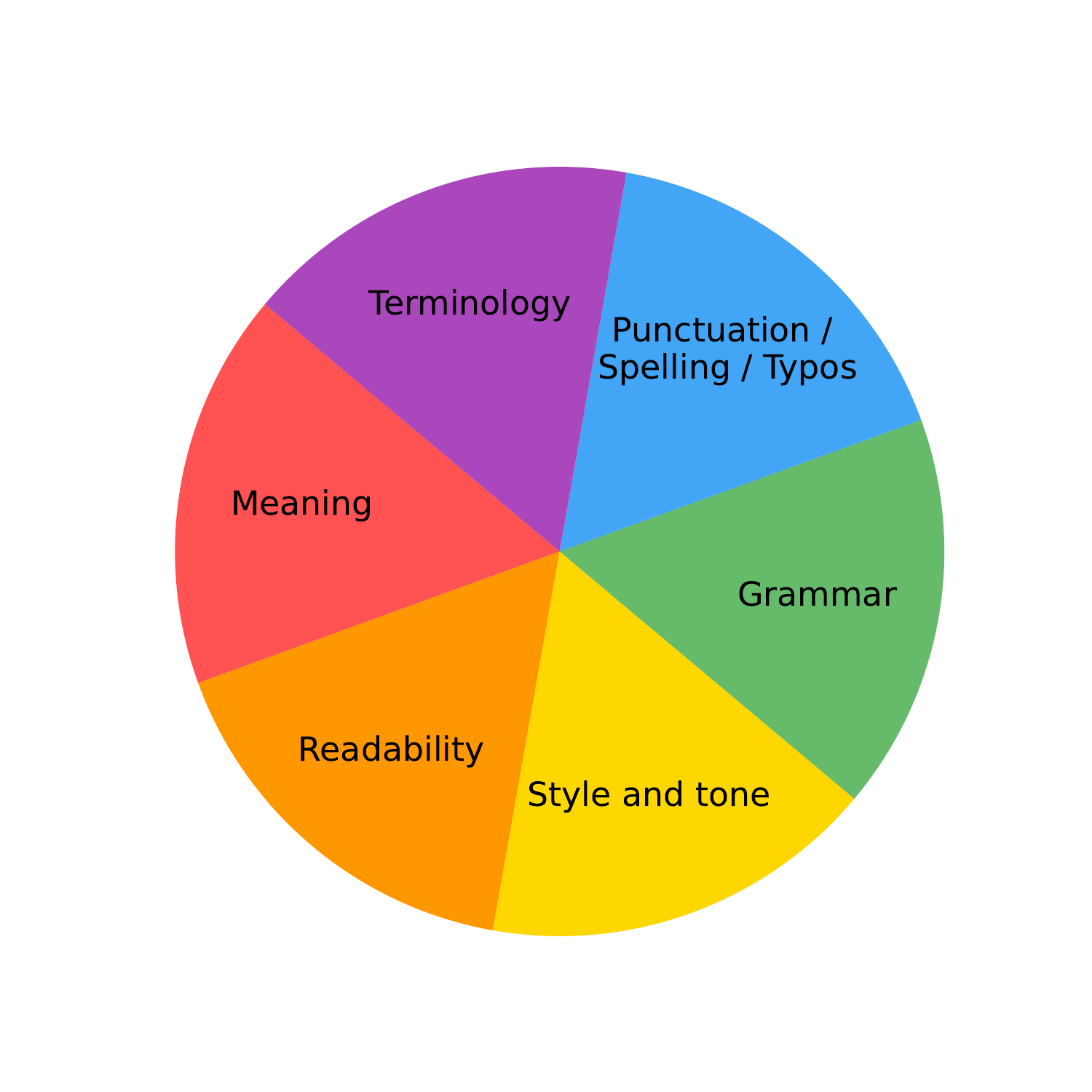} 
    \end{minipage}%
    \caption{(Left) Translation Quality Assessment. Auto-MQM scores $\ge 0.80$ are considered professional quality, and Human Eval percentages represent the rate of accurate translations. (Right) Specific dimensions that are considered in Auto-MQM evaluation.}
    \label{fig:translation_eval}
\end{figure}

In order to optimize quality of translations in the benchmark, we implemented a dual-verification process comprising Multidimensional Quality Metrics (MQM) and a meta human review.
The Auto-MQM framework evaluates translations across multiple issue dimensions, as illustrated in the pie chart in Figure~\ref{fig:translation_eval} (right). 
Complementing this, the meta human review serves as the human validation layer where native experts evaluated translation accuracy by sampling 5\% to 10\% of unique strings for each task across all languages.

The evaluation results are presented in the table in Figure~\ref{fig:translation_eval} (left).
Under the Auto-MQM framework, translations across all tasks achieved professional quality standards.
Due to WMT24++ and IndoMMLU exceeding our automated quality thresholds, we exempted these tasks from the meta review.
For the remaining tasks, the human evaluation accuracy is consistent with the automated metrics, which further validates the fidelity of the newly generated data in \dataset{} benchmark.

\begin{table}[t]
\centering
\resizebox{0.5\textwidth}{!}{%
\begin{tabular}{lll}
\toprule
\textbf{Dataset} & \textbf{Existing} & \textbf{Newly Created} \\
\midrule
XQuAD & vi, th & id, fil, ceb, ilo, jv, km, lo, ms, su, my \\
CrossSum & th, vi, id, my & fil, ceb, ilo, jv, km, lo, ms, su \\
IndoMMLU & id & th, vi, fil, ceb, ilo, jv, km, lo, ms, su, my \\
MGSM & th & vi, id, fil, ceb, ilo, jv, km, lo, ms, su, my \\
WMT24++ & vi, th, id, fil & ceb, ilo, jv, km, lo, ms, su, my \\
\bottomrule
\end{tabular}}
\caption{Overview of the data created and reused from existing collections, showing ISO 639-1 language codes.}
\label{tab:dataset-availability}
\end{table}

\begin{table*}[t]
\centering
\resizebox{1.0\textwidth}{!}{
\begin{tabular}{@{}ccccccccccccccc@{}}
\toprule
\multirow{2}{*}{\textbf{Task}} & \multirow{2}{*}{\textbf{Split}} & \multicolumn{13}{c}{\textbf{Language (code)}} \\
& & \begin{tabular}[c]{@{}c@{}}English\\ (en)\end{tabular} & \begin{tabular}[c]{@{}c@{}}Thai\\ (th)\end{tabular} & \begin{tabular}[c]{@{}c@{}}Vietnamese\\ (vi)\end{tabular} & \begin{tabular}[c]{@{}c@{}}Indonesian\\ (id)\end{tabular} & \begin{tabular}[c]{@{}c@{}}Filipino\\ (fil)\end{tabular} & \begin{tabular}[c]{@{}c@{}}Malay\\ (ms)\end{tabular} & \begin{tabular}[c]{@{}c@{}}Cebuano\\ (ceb)\end{tabular} & \begin{tabular}[c]{@{}c@{}}Ilocano\\ (ilo)\end{tabular} & \begin{tabular}[c]{@{}c@{}}Javanese\\ (jv)\end{tabular} & \begin{tabular}[c]{@{}c@{}}Sundanese\\ (su)\end{tabular} & \begin{tabular}[c]{@{}c@{}}Khmer\\ (km)\end{tabular} & \begin{tabular}[c]{@{}c@{}}Lao\\ (lo)\end{tabular} & \begin{tabular}[c]{@{}c@{}}Burmese\\ (my)\end{tabular} \\ \midrule
\multirow{3}{*}{\begin{tabular}[c]{@{}c@{}}Mathematical Reasoning\\ (MGSM)\end{tabular}} & test & 250 & 250 & 250 & 250 & 250 & 250 & 250 & 250 & 250 & 250 & 250 & 250 & 250 \\
& dev & 8 & 8 & 8 & 8 & 8 & 8 & 8 & 8 & 8 & 8 & 8 & 8 & 8 \\
& train & - & - & - & - & - & - & - & - & - & - & - & - & - \\ \midrule
\multirow{3}{*}{\begin{tabular}[c]{@{}c@{}}Scientific Reasoning\\ (IndoMMLU)\end{tabular}} & test & 735 & 735 & 735 & 735 & 735 & 735 & 735 & 735 & 735 & 735 & 735 & 735 & 735 \\
& dev & 15 & 15 & 15 & 15 & 15 & 15 & 15 & 15 & 15 & 15 & 15 & 15 & 15 \\
& train & - & - & - & - & - & - & - & - & - & - & - & - & - \\ \midrule
\multirow{3}{*}{\begin{tabular}[c]{@{}c@{}}Question Answering\\ (XQuAD)\end{tabular}} & test & 1190 & 1190 & 1190 & 1190 & 1190 & 1190 & 1190 & 1190 & 1190 & 1190 & 1190 & 1190 & 1190 \\
& dev & 111 & 111 & 111 & 111 & 111 & 111 & 111 & 111 & 111 & 111 & 111 & 111 & 111 \\
& train & 103 & 103 & 103 & 103 & 103 & 103 & 103 & 103 & 103 & 103 & 103 & 103 & 103 \\ \midrule
\multirow{3}{*}{\begin{tabular}[c]{@{}c@{}}Abstractive Summarization\\ (CrossSum)\end{tabular}} & test & 500 & 101 & 457 & 500 & 500 & 500 & 500 & 500 & 500 & 500 & 500 & 500 & 69 \\
& dev & 100 & 100 & 100 & 100 & 100 & 100 & 100 & 100 & 100 & 100 & 100 & 100 & 74 \\
& train & 100 & 100 & 100 & 100 & 100 & 100 & 100 & 100 & 100 & 100 & 100 & 100 & 100 \\ \midrule
\multirow{3}{*}{\begin{tabular}[c]{@{}c@{}}Machine Translation\\ (WMT24++)\end{tabular}} & test & - & 997 & 997 & 997 & 997 & 997 & 997 & 997 & 997 & 997 & 997 & 997 & 997 \\
& dev & - & - & - & - & - & - & - & - & - & - & - & - & - \\
& train & - & - & - & - & - & - & - & - & - & - & - & - & - \\ \bottomrule
\end{tabular}
}
\caption{Data statistics in \dataset{} benchmark. The table details the number of instances for each task, broken down by language and data split (test, dev, and train). Note that for the CrossSum task, the splits of th, id, vi, and my languages do not have a one-to-one mapping from the English data.}
\label{tab:statistics_all}
\end{table*}



\begin{table*}[ht]
\centering
\resizebox{\textwidth}{!}{%
\begin{tabular}{@{} l *{14}{wc{0.9cm}} @{}}
\toprule
\multirow{2.5}{*}{\textbf{Model}}
& \multicolumn{3}{c}{\textbf{\makecell{Mathematical\\Reasoning}}}
& \multicolumn{3}{c}{\textbf{\makecell{Scientific\\Reasoning}}}
& \multicolumn{3}{c}{\textbf{\makecell{Question\\Answering}}}
& \multicolumn{3}{c}{\textbf{\makecell{Abstractive\\Summarisation}}}
& \multicolumn{2}{c}{\textbf{\makecell{Machine\\Translation}}} \\
\cmidrule(lr){2-4}
\cmidrule(lr){5-7}
\cmidrule(lr){8-10}
\cmidrule(lr){11-13}
\cmidrule(lr){14-15}
& \multicolumn{3}{c}{Accuracy $\uparrow$}
& \multicolumn{3}{c}{Accuracy $\uparrow$}
& \multicolumn{3}{c}{F1 $\uparrow$}
& \multicolumn{3}{c}{ChrF $\uparrow$}
& \multicolumn{2}{c}{MetricX-24 $\downarrow$} \\
\cmidrule(lr){2-4}
\cmidrule(lr){5-7}
\cmidrule(lr){8-10}
\cmidrule(lr){11-13}
\cmidrule(lr){14-15}
\textbf{Language}
& en & high & low
& en & high & low
& en & high & low
& en & high & low
& high & low \\
\midrule
Gemini-2.5-Pro
& 97.2 & 94.5 & 94.0
& 84.2 & 83.8 & 83.5
& 83.7 & 79.8 & 75.6
& 35.8 & 34.7 & 34.2
& 2.6 & 3.7 \\
Gemini-2.5-Flash
& 97.2 & 92.9 & 92.0
& 83.7 & 82.4 & 82.2
& 82.9 & 78.2 & 74.7
& 33.7 & 34.0 & 34.1
& 2.7 & 3.8 \\
\midrule
Gemma-2-27B
& 93.2 & 87.9 & 71.4
& 67.5 & 64.3 & 57.1
& 85.2 & 76.3 & 69.2
& \underline{35.0} & 32.7 & 27.7
& 3.4 & 9.3 \\
Gemma-2-9B
& 91.2 & 81.9 & 61.2
& 65.2 & 61.1 & 53.2
& 79.4 & 61.7 & 57.1
& 33.5 & 30.4 & 21.2
& 3.9 & 11.6 \\
SEA-LION-v3-9B
& 94.0 & 85.0 & 70.0
& 71.2 & 66.0 & 57.1
& 83.6 & 72.7 & 65.2
& 33.1 & 32.4 & 27.2
& 3.4 & 9.3 \\
\bottomrule
\end{tabular}
}
\caption{Performance scores of additional models on \dataset{}, which are divided into English, high- and low-resource non-English language groups.}
\label{tab:new_model_comparison}
\end{table*}

\section{Performance Comparison in Zero-shot and Few-shot Prompting}
\begin{figure*}[t]
    \centering
    \includegraphics[width=1.0\linewidth]{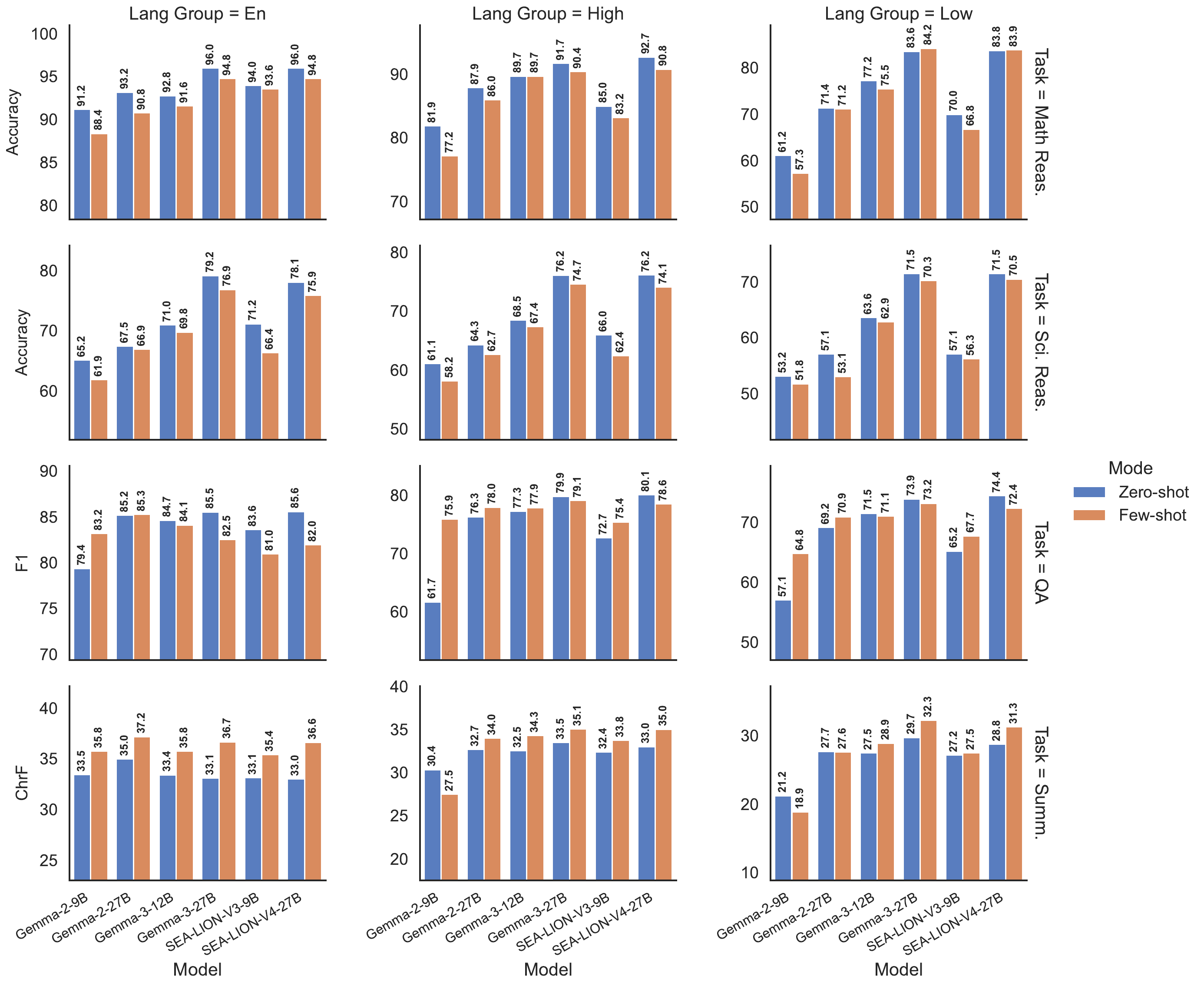}
    \caption{Comparison of various models in zero-shot and few-shot (three-shot) settings.}
    \label{fig:fs_result}
\end{figure*}

To assess the effectiveness of in-context learning across different model architectures and language groups, we conducted a comparative analysis of zero-shot and few-shot (three-shot) prompting on mathematical reasoning, scientific reasoning, question answering (QA), and abstractive summarization.
The results are illustrated in Figure~\ref{fig:fs_result}.

Our results reveal a distinct divergence based on task type.
On reasoning-heavy tasks (mathematical and scientific reasoning), the few-shot performance is notably worse than zero-shot.
We hypothesize that this degradation occurs because the few-shot exemplars provided only direct answers without the intermediate reasoning steps; this likely constrained the models to a direct-answer format, whereas the zero-shot setting better allowed the models to inherently generate the necessary chain-of-thought.
Conversely, for the generative tasks (QA and summarization), the few-shot setting proved significantly more effective on most models, likely because the exemplars provided necessary structural guidance and aligned the models with the expected output format helping to prime the model with essential multilingual context to better activate the model's latent cross-lingual knowledge.

\clearpage
\onecolumn
\section{Prompts Used In Experiments}
\label{appd:prompts}

The first five prompts below are the default zero-shot prompts used for evaluating the five benchmark tasks. The final three prompts are used specifically for the analysis experiments: the English CoT variants for MGSM and IndoMMLU, and the English-pivot variant for CrossSum. For brevity, we present only the English versions of the reasoning-language prompt templates as representative examples. In the corresponding Native CoT condition,  the entire prompt was translated into each target language, and models were explicitly instructed to provide their reasoning in that language.

\begin{promptbox}{Zero-shot Prompt for MGSM}
You are an expert in solving math reasoning problems. Please solve the following math word problem.
The problem is presented in its original language. Please read it carefully. Please provide your step-by-step reasoning and solution in the SAME language as the problem. Please conclude with the final answer in the format: The final answer is \textless answer\textgreater NUMERICAL\_ANSWER\textless /answer\textgreater
\\ \\
\{problem\}
\end{promptbox}

\begin{promptbox}{Zero-shot Prompt for IndoMMLU}
You are an expert in solving scientific reasoning problems. Please solve the following multiple choice problem.
The problem is presented in its original language. Please read it carefully.
Please provide your thinking process in the SAME language as the problem.
Please conclude with the final answer in the format: The final answer is \textless answer\textgreater OPTION\textless /answer\textgreater
\\ \\
The following are multiple choice questions (with answers) about \{subject\}.
\\ \\
Question: \{question\} \\
A. \{option\_A\} \\
B. \{option\_B\} \\
C. \{option\_C\} \\
D. \{option\_D\}
\end{promptbox}

\begin{promptbox}{Zero-shot Prompt for CrossSum}
You are an expert multilingual assistant specializing in text summarization.
Your task is to read the following document written in English and generate a concise summary of it in target language.
The summary MUST be only ONE sentence and capture the absolute main idea or most significant takeaway of the document. The summary must be factually consistent with the source document.
\\ \\
English Document: \\
\{document\}
\\ \\
Please provide the ONE-SENTENCE summary in \{target\_language\} below:
\\ \\
\{target\_language\} Summary:
\end{promptbox}

\begin{promptbox}{Zero-shot Prompt for XQuAD}
You are an expert in solving question answering problems. Please solve the following problem.
Given the context and question, you should answer the question by extracting the answer *directly* from the context.
The context and question are presented in its original language. Please read it carefully.
Please provide your thinking process and answer in the SAME language as the problem.
Please conclude with the final answer in the format: The final answer is \textless answer\textgreater ANSWER\textless /answer\textgreater
\\ \\
Context: \\
\{context\}
\\ \\
Question: \\
\{question\}
\end{promptbox}

\begin{promptbox}{Zero-shot Prompt for WMT24++}
You are a professional Southeast Asia language translator, tasked with providing translations suitable for use in Southeast Asia region.
Your goal is to accurately convey the meaning and nuances of the original English text while adhering to target language grammar, vocabulary, and cultural sensitivities.
Produce only the target translation, without any additional explanations or commentary.
\\ \\
Please translate the following English text into \{target\_language\}: \\
\{input\_text\}
\\ \\
\{target\_language\} Translation:
\end{promptbox}

\begin{promptbox}{Zero-shot Prompt for MGSM with English Reasoning}
You are an expert in solving math reasoning problems. Please solve the following math word problem.
The problem is presented in its original language. Please read it carefully. Please provide your step-by-step reasoning and solution in \textbf{English}. Please conclude with the final answer in the format: The final answer is \textless answer\textgreater NUMERICAL\_ANSWER\textless /answer\textgreater
\\ \\
\{problem\}
\end{promptbox}

\begin{promptbox}{Zero-shot Prompt for IndoMMLU with English Reasoning}
You are an expert in solving scientific reasoning problems. Please solve the following multiple choice problem.
The problem is presented in its original language. Please read it carefully.
Please provide your thinking process in \textbf{English}.
Please conclude with the final answer in the format: The final answer is \textless answer\textgreater OPTION\textless /answer\textgreater
\\ \\
The following are multiple choice questions (with answers) about \{subject\}.
\\ \\
Question: \{question\} \\
A. \{option\_A\} \\
B. \{option\_B\} \\
C. \{option\_C\} \\
D. \{option\_D\}
\end{promptbox}

\begin{promptbox}{Zero-shot Prompt for CrossSum with English Pivot}
You are an expert multilingual assistant specializing in text summarization.
Your task is to read the following document written in English and generate a concise summary of it.
The summary MUST be only ONE sentence and capture the absolute main idea or most significant takeaway of the document. The summary must be factually consistent with the source document.
You should first generate the summary in English, then translate it into target language.
\\ \\
You MUST follow this template EXACTLY. Do not add any extra text outside the tags.
\\ \\
\textless English\_Summary\textgreater \\
{[}A concise, single-sentence summary of the document in English{]} \\
\textless /English\_Summary\textgreater \\
\textless Target\_Summary\textgreater \\
{[}The translation of the single-sentence summary into target language{]} \\
\textless /Target\_Summary\textgreater
\\ \\
Target Language: \{target\_language\}
\\ \\
\textless Source\_Document\textgreater \\
\{document\} \\
\textless /Source\_Document\textgreater
\end{promptbox}